\documentclass{article}

\usepackage[preprint,nonatbib]{neurips_2025}

\usepackage{cite}
\usepackage[utf8]{inputenc} 
\usepackage[T1]{fontenc}    
\usepackage{hyperref}       
\usepackage{url}            
\usepackage{booktabs}       
\usepackage{amsfonts}       
\usepackage{nicefrac}       
\usepackage{microtype}      
\usepackage{xcolor}         
\usepackage{graphicx}
\usepackage{float}
\usepackage{hhline}
\usepackage{amsmath}

\usepackage{caption}
\usepackage{verbatim}
\usepackage[T1]{fontenc}
\usepackage{multirow}
\usepackage{tabularx}
\usepackage{makecell}
\usepackage{marvosym}
\usepackage{arydshln}
\usepackage{enumitem}

\title{MedAgent-Pro: Towards Evidence-based Multi-modal Medical Diagnosis via Reasoning Agentic Workflow}

\author{%
    Ziyue Wang$^{1}$, 
    Junde Wu$^{2}$, 
    Linghan Cai$^{3}$,
    Chang Han Low$^{1}$, \\
    \textbf{Xihong Yang$^{1,4}$},
    \textbf{Qiaxuan Li$^{5}$},
    \textbf{Yueming Jin$^{1}$\thanks{Corresponding author: ymjin@nus.edu.sg}}\\
    $^{1}$National University of Singapore,\quad
    $^{2}$University of Oxford, \\
    $^{3}$Harbin Institute of Technology (Shenzhen) \quad
    $^{4}$National University of Defense Technology\\
    $^{5}$The Second Affiliated Hospital Zhejiang University School of Medicine 
}

\begin{document}

\maketitle

\begin{abstract}
In modern medicine, clinical diagnosis relies on the comprehensive analysis of primarily textual and visual data, drawing on medical expertise to ensure systematic and rigorous reasoning.
Recent advances in large Vision–Language Models (VLMs) and agent-based methods hold great potential for medical diagnosis, thanks to the ability to effectively integrate multi-modal patient data. However, they often provide direct answers and draw empirical-driven conclusions without quantitative analysis, which reduces their reliability and clinical usability.
We propose MedAgent-Pro, a new agentic reasoning paradigm that follows the diagnosis principle in modern medicine, to decouple the process into sequential components for step-by-step, evidence-based reasoning.
Our MedAgent-Pro workflow presents a hierarchical diagnostic structure to mirror this principle, consisting of disease-level standardized plan generation and patient-level personalized step-by-step reasoning.
To support disease-level planning, an RAG-based agent is designed to retrieve medical guidelines to ensure alignment with clinical standards. For patient-level reasoning, we propose to integrate professional tools such as visual models to enable quantitative assessments. Meanwhile, we propose to verify the reliability of each step to achieve evidence-based diagnosis, enforcing rigorous logical reasoning and a well-founded conclusion.
Extensive experiments across a wide range of anatomical regions, imaging modalities, and diseases demonstrate the superiority of MedAgent-Pro to mainstream VLMs, agentic systems and state-of-the-art expert models. 
Ablation studies and human evaluation by clinical experts further validate its robustness and clinical relevance.
Code is available at https://github.com/jinlab-imvr/MedAgent-Pro.
\end{abstract}

\begin{figure}[ht]
    \centering
    \includegraphics[width=0.83\linewidth]{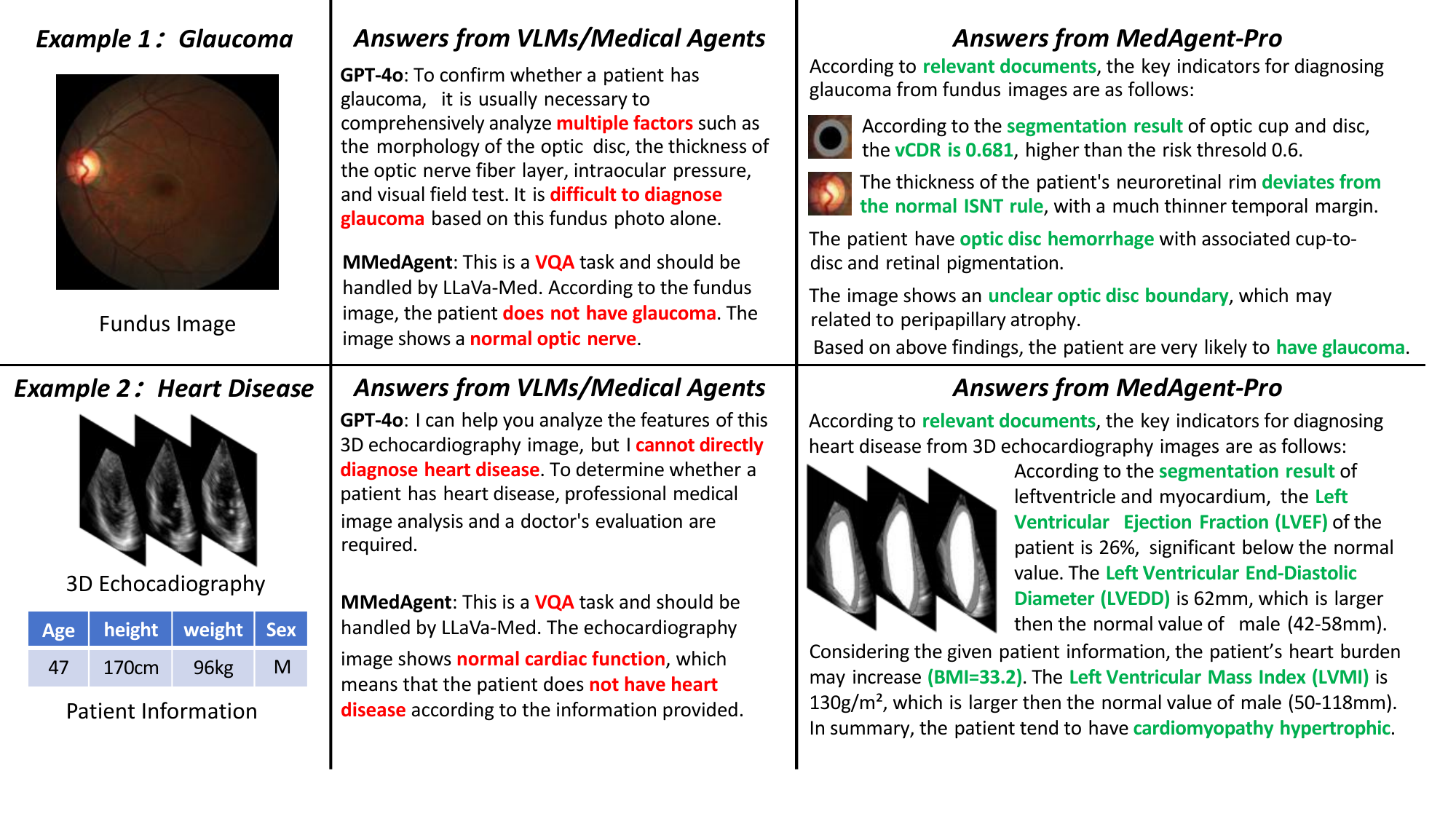}
    \caption{Comparison of diagnostic outcomes for two diseases across mainstream VLMs, medical agentic systems, and our proposed MedAgent-Pro workflow.
    }
    \label{fig:effect}
    \vspace{-15pt}
\end{figure}

\vspace{-11pt}
\section{Introduction}
\vspace{-7pt}
Clinical diagnosis, a core task in medical practice, entails synthesizing various clinical information to reach a conclusion \cite{guideline,evi1,ev2}, where clinicians make decisions mainly based on visual cues from medical imaging and textual information from patient records.
For example, it is common for clinicians to examine radiology images to identify tumor scales, or analyze pathological images to detect potential cancer.
Early AI-assisted methods focus on adopting pure image analysis models to support diagnosis, such as tumor segmentation in oncology and pneumonia severity assessment in chest radiology. 
Recently, Vision-Language Models (VLMs) have demonstrated that integrating multi-modal information can significantly benefit diagnosis, and medical visual question answering (VQA) is a key task in this context, where models answer textual questions based on medical images \cite{llavamed,vqa1,vqa_rad,pathvqa}.

However, ordinary VQA formulated in the existing methods still fails to reflect the real-world diagnostic processes. 
Rather than performing such one-hop QA, clinical diagnosis involves a standardized, step-by-step process \cite{guideline,ev4,ev5,ev6,ev7}. 
The diagnostic process for each patient typically involves two stages:
i) Determining the target disease and formulating a standardized workflow based on medical guidelines with multiple clinical indicators to support the decision;
ii) Step-by-step analysis of personalized data, combining qualitative and quantitative assessments to evaluate these indicators.
Throughout the process, each analytical step builds on previous assumptions and is supported by relevant literature or analysis provided by specialized tools
For example, in glaucoma diagnosis, the cup-to-disc ratio is a key indicator, which relies on the accurate localization of the optic cup and disc regions in the preceding step.
In contrast, conventional VQA models often generate diagnostic conclusions hastily, relying on empirical internal knowledge without fine-grained analysis.

Due to the strict requirement of medical diagnosis, current methods fall short of meeting the clinical standard:
VLMs \cite{gpt4, janus, deepseek, llavamed, llava} have shown strong performance across a range of tasks \cite{medflamingo, pathfinder, clinicaldecisionsupport}, however, they lack sufficient medical knowledge and remain inadequate for in-depth medical analysis. 
While models like GPT-o1 \cite{gpt4} and DeepSeek-R1 \cite{deepseek} incorporate reasoning capabilities to support more structured medical analysis, their limited fine-grained visual perception ability impairs quantitative analysis and reduces their effectiveness in clinical applications.
Meanwhile, agentic systems \cite{autoagents, camel, reflexion, proagent, ltc} have extended the capabilities of VLMs by introducing more professional models \cite{biomedclip, maria, medsam, msa, medsam2}. 
However, current medical agentic systems \cite{mdagent, medagents, kg4diagnosis, mmedagent, medrax} simply glued all tools together instead of a clinically oriented workflow, functioning more as an integrated toolbox than an end-to-end automated solution.
As a result, when asked to provide a diagnosis, these systems simply invoke the internal VLM without selecting appropriate tools to support their decision-making.
In summary, existing methods treat medical diagnosis as an empirical one-hop question-answering task, relying solely on VLMs' internal knowledge to make qualitative judgments. However, modern medical practice emphasizes evidence-based diagnosis, which requires structured reasoning and clinical evidence \cite{ev4,ev5,ev6,ev7}.

To tackle these issues, we propose MedAgent-Pro, a reasoning agentic workflow tackling versatile multi-modal medical diagnosis tasks. We aim to design a workflow that aligns with modern medical criteria, provides decision support with medical guidelines and quantitative analysis as shown in Fig.~\ref{fig:effect}.
Our MedAgent-Pro embraces a hierarchical structure to simulate the modern clinical procedure, conduct step-by-step reasoning at the disease and patient levels. \textit{Disease-level planning} generates standardized diagnostic plans, while \textit{Patient-level reasoning} follows these plans to analyze personalized information.
To ensure that diagnostic plans align with clinical guidelines, MedAgent-Pro incorporates an RAG agent that retrieves relevant medical knowledge. In line with modern clinical workflows, which routinely employ specialized tools for diagnostic support, MedAgent-Pro integrates expert tools such as visual models to enable accurate quantitative evaluation of clinical indicators. Furthermore, to maintain the rigor of multi-step clinical reasoning, we propose an evidence-based analysis in which the system evaluates the reliability of each step’s output before proceeding to the next step, ensuring that every diagnostic inference is grounded in a sound and trustworthy foundation.
Our key contributions are summarized as:
\begin{itemize}[leftmargin=*,labelsep=0.5em]
\item We propose MedAgent-Pro, the first agentic paradigm that presents systematic, evidence-based reasoning for accurate and reliable medical diagnosis. By aligning with the principle of modern medical workflow, our paradigm transforms the empirical, ready-made outputs of prior methods into more rigorous logical reasoning and a well-founded conclusion.
\item MedAgent-Pro presents a hierarchical structure consisting of \textit{disease-level} and \textit{patient-level} reasoning. A RAG-based method ensures that disease-level planning aligns with medical guidelines, while quantitative analysis and evidence-based reasoning are devised at the patient level to ensure professionalism and reliability of the step-by-step analysis.
\item We validate MedAgent-Pro comprehensively across 10+ imaging modalities, 20+ anatomies, and 50+ diseases, surpassing mainstream VLMs, medical agentic systems, and even task-specific models.
Notably, it outperforms GPT-4o by 34\% and 22\% on glaucoma and heart disease diagnosis. 
Clinician evaluations further highlight diagnostic quality and reliability of our MedAgent-Pro.
\end{itemize}

\vspace{-10pt}
\section{Related work}
\vspace{-5pt}
\paragraph{Multi-modal Medical Diagnosis}
Developing AI techniques for multi-modal medical diagnosis has become a primary research objective. \cite{diagnosis3, diagnosis1, diagnosis2, diagnosis4}.
Prior research focused on medical imaging assessment, including classification \cite{biomedclip, classification1, classification2, classification3, classification4}, detection \cite{detection1, detection2, detection3, detection4, detection5}, and segmentation \cite{unet, nnunet, donuseg, cell1, msa, medsam, dawn,seine}.
VQA has been proposed for end-to-end multi-modal diagnosis 
\cite{pmc_vqa, mmbert, vqa1}, while VLMs \cite{llavamed, Med-flamingo,medfilip, ctglip} have yielded competitive performance in medical VQA. 
Despite these advancements, medical VQA \cite{vqa_rad, slake, pathvqa} remain overly simplified compared to the diagnostic practice,
and further research is highly desired.
\vspace{-5pt}
\paragraph{VLM-based AI Agent}
Developing autonomous intelligent systems is a long-standing research goal, with agent-based methods gaining increasing attention.
VLM-based agents have made significant advancements in diverse applications such as industrial engineering \cite{a1, a2, a3, guanyi}, scientific experimentation \cite{a4, a5, a6}, embodied agents \cite{a7, a8, a9}, gaming \cite{a10,a11,a12}, and societal simulation \cite{a14, a15, visionzip}.
Despite their adaptability across diverse applications without additional training, VLM-based agents remain limited in the medical domain due to insufficient fine-grained visual perception.
\vspace{-5pt}
 \paragraph{Medical Agentic System}
 Current medical agentic systems can be categorized into two streams. The first stream \cite{mdagent,medagents,kg4diagnosis} enhances VLM capabilities by incorporating mechanisms such as debate or majority voting among multiple VLMs to refine responses. The second stream \cite{mmedagent, medrax} functions as a toolkit for diverse medical tasks, integrating an orchestrator agent with various specialized models. However, they simply glued all tools together instead of a clinically oriented workflow, functioning more as an integrated toolbox, and cannot handle complex diagnoses.

\begin{figure}[b]
    \centering
    \includegraphics[width=0.95\linewidth]{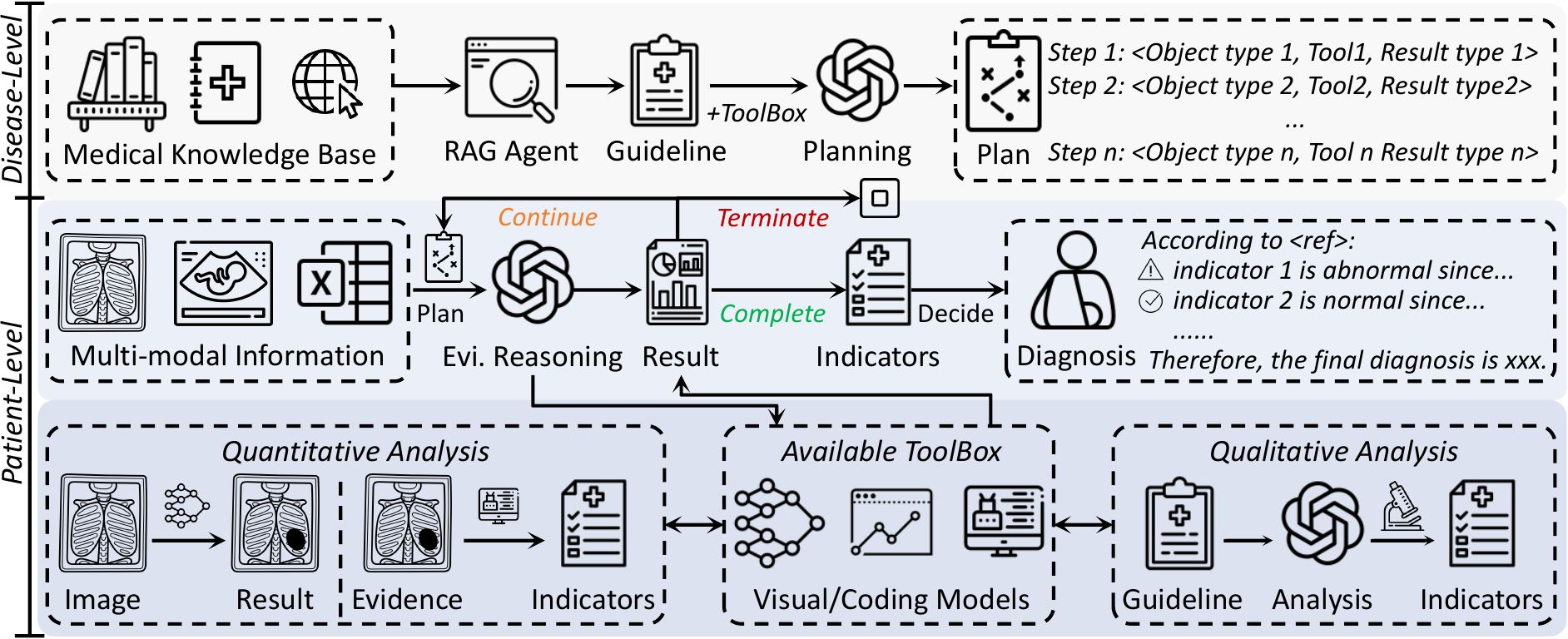}
    \caption{Overview of the MedAgent-Pro framework, which performs diagnosis through a hierarchical structure, with reasoning guided by a VLM supported by an RAG agent and specialized tools.}
    \label{fig:enter-stru}
\end{figure}

\vspace{-5pt}
\section{Methods}
\vspace{-5pt}
\subsection{Overall Workflow}
\vspace{-5pt}
For a specific disease, clinicians typically develop a standardized diagnostic workflow based on medical guidelines. Each patient's diagnosis then follows this workflow, combining qualitative observations with quantitative assessments to obtain clinical indicators \cite{evi1,ev2,ev3,ev4}.
To mirror this clinical paradigm, we design a hierarchical reasoning workflow consisting of \textit{disease-level planning} and \textit{patient-level reasoning } in MedAgent-Pro, to enable personalized diagnosis for each patient under standardized, disease-specific guidance. During the workflow, we take a VLM $\mathcal{V}$ as a baseline model to conduct basic tasks within the workflow.
As illustrated in Fig.~\ref{fig:enter-stru}, the \textit{disease-level planning} is designed to formulate standardized diagnostic plans for each disease based on medical guidelines, which is assisted by an RAG agent. 
Meanwhile, \textit{patient-level reasoning} processes each patient's personalized data individually by executing the diagnostic plan step by step and assisting the VLM in performing necessary quantitative analysis through specialized tools, thereby enabling evidence-based reasoning.
We provide a detailed explanation in the following sections.

\subsection{Disease-level Knowledge-based Planning}
In clinical practice, doctors develop standardized workflows for specific diseases based on their expertise and experience \cite{ev5,ev6,ev7,ev8}. 
Following this routine, we introduce a Retrieval-Augmented Generation (RAG) agent $\mathcal{R}$ to incorporate medical guidelines during the planning stage to guide diagnostic plan generation. 
MedAgent-Pro is equipped with a domain-specific knowledge base $\mathcal{K}$, built from MedlinePlus \cite{national2006medlineplus, miller2000medlineplus}, which includes entries on over 1,000 diseases and conditions, and more than 4,000 expert-reviewed articles on symptoms, tests, injuries, and treatments.
\begin{figure}[h]
    \centering
    \includegraphics[width=0.98\linewidth]{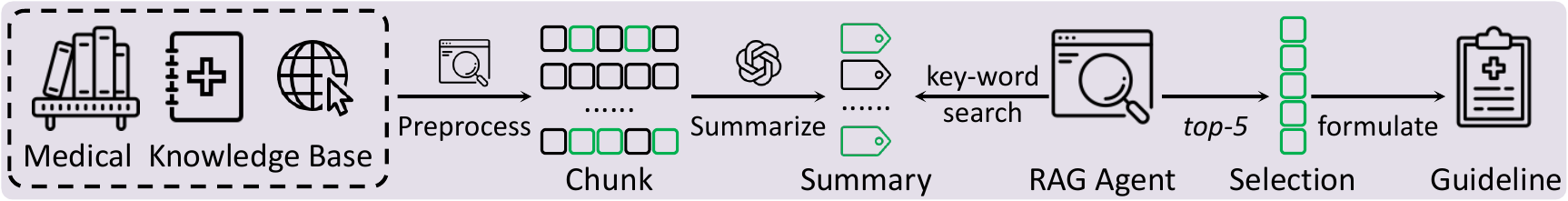}
    \caption{The illustration of the RAG process, which leverages a two-step retrieval.}
    \label{fig:rag}
\end{figure}
As shown in Fig.~\ref{fig:rag}, the knowledge base $\mathcal{K}$ is first indexed into a vector database, where each document is split into chunks. To accelerate retrieval, a one-sentence summary is pre-generated for each document and used as its index. Upon receiving a disease query, the RAG agent $\mathcal{R}$ filters out irrelevant entries via keyword search over these summaries, then conducts vector retrieval within the remaining documents to extract the top 5 most relevant chunks.
Based on the retrieved chunks, the VLM summarizes their content and generates a procedural guideline $\mathcal{G}$ that reflects real-world clinical practices for the queried disease. The generation process of $\mathcal{G}$ can be formulated as:
\begin{equation}
    \mathcal{G} = \mathcal{V}(\mathcal{R} (\mathcal{K})).
\end{equation}
The VLM first summarizes a set of disease-specific clinical indicators $\mathcal{I} = \{\mathcal{I}_1, \mathcal{I}_2, \dots, \mathcal{I}_m\}$ from the guideline $\mathcal{G}$. 
To support the analysis of $\mathcal{I}$, MedAgent-Pro is equipped with a toolset $\mathcal{T}$.
To construct an executable plan, the VLM integrates $\mathcal{G}$ with a predefined set of operation descriptions $\mathcal{A}$, where each element $a \in \mathcal{A}$ corresponds one-to-one with a tool $t \in \mathcal{T}$ (i.e., $t$ can be a segmentation model, and $a$ is its paired description, such as: "This model segments the optic cup in a fundus image.") 
Based on $\mathcal{G}$ and $\mathcal{A}$, the VLM generates a disease-specific diagnostic plan comprising multiple reasoning steps $\mathcal{P} = \{\mathcal{P}_1, \mathcal{P}_2, \dots, \mathcal{P}_n\}$.
Each step $\mathcal{P}_i \in \mathcal{P}$ is represented as:
\begin{equation}
\mathcal{P}_i:
r_i = \langle k_{r_i}, v_{r_i} \rangle = \langle k_{r_i}, t(v_{o_i}) \rangle, t \in \mathcal{T}
\end{equation}
where the processed object $o_i = \langle k_{o_i}, v_{o_i} \rangle$ and result $r_i = \langle k_r, v_r \rangle$ are represented as key-value pairs, the key ($k_{o_i}$ or $k_{r_i}$) specifies the data property, and the value ($v_{o_i}$ or $v_{r_i}$) contains the actual data.

In practice, $\mathcal{P}$ is stored as a JSON file. Each $\mathcal{P}_i$ includes an operation key $t$, a predefined Python function from the toolset $\mathcal{T}$ with fixed input-output behavior, and two data fields $k_o$ and $k_r$ specifying the expected input and output data property for $t$. During reasoning, the VLM checks the data property of the current input. If it matches any $k_o$ in the JSON entry, the corresponding function $t$ is invoked to obtain the output data property $k_r$. 
For example, when provided with a fundus image, MedAgent-Pro performs the designated operation and generates the corresponding optic cup segmentation mask.
Throughout the design, each disease is assigned a diagnosis plan aligned with medical guidelines, enabling a regulated and standardized workflow for every patient.

\begin{figure}[t]
    \centering
    \includegraphics[width=0.9\linewidth]{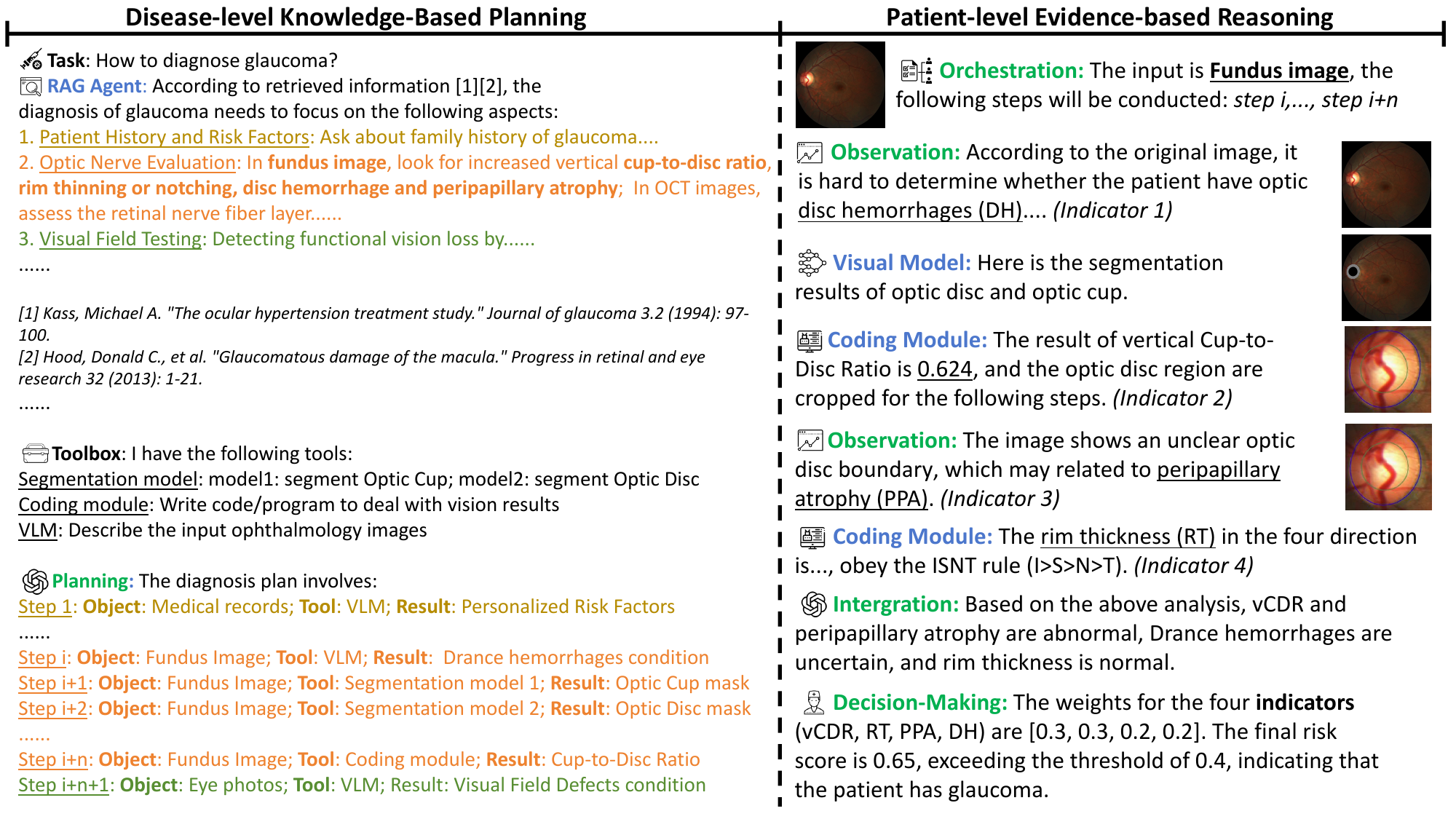}
    \caption{A case study for glaucoma diagnosis, which presents a detailed workflow in the MedAgent-Pro framework. The blue text indicates the agents, while the green text indicates the reasoning steps.
    In the reasoning, the underlined text indicates the clinical indicators identified through analysis.
    }
    \label{fig:case}
    \vspace{-10pt}
\end{figure}

\subsection{Patient-level Evidence-based Reasoning}
For each patient case of the disease, MedAgent-Pro follows the generated disease-specific diagnostic plan $\mathcal{P}$ to analyze clinical indicators $\mathcal{I}$. 
For each patient's personalized multi-modal data $\mathcal{D}$, the VLM performs an orchestration process to select executable steps from $\mathcal{P}$ based on data availability, filtering out those requiring unavailable inputs.
\begin{equation}
\mathcal{P}' = \mathcal{V}(\mathcal{P}, \mathcal{D}) \quad \text{s.t. } \forall \mathcal{P}_i = \langle k_{r_i}, v_{r_i} \rangle = \langle k_{r_i}, t(v_{o_i}) \rangle, ;  k_{o_i} \in \mathcal{D}.
\end{equation}
As shown in Fig.~\ref{fig:case}, in the glaucoma diagnosis, when provided with the fundus image, the orchestration process selects relevant steps and skips those requiring unavailable data such as OCT or visual field. 
A core principle of modern medicine is evidence-based practice, which emphasizes the conscientious and judicious use of reliable clinical evidence, integrated with individual clinical expertise, to guide patient care decisions \cite{evi1,ev2,ev3,ev4}.
Following this principle, we invoke specialized tools for quantitative assessments and adopt an evidence-based reasoning paradigm to ensure the reliability.

\paragraph{Quantitative Analysis}
For quantitative analysis, professional tool agents are engaged to perform specialized assessments to bridge the gap between AI and clinical practice. 
The toolset $\mathcal{T}$ includes visual models such as segmentation tools (e.g., Medical SAM Adapter \cite{msa}, MedSAM \cite{medsam}, Cellpose \cite{cellpose1}), grounding models (e.g., Maira-2 \cite{maria}), and LLM-based coding tools (e.g., Copilot \cite{web:copilot}).
As shown in Fig.~\ref{fig:case},  specialized segmentation tools are used to extract the optic cup and disc masks, while coding tools subsequently compute the cup-to-disc ratio based on the segmentation results, which serves as a key indicator in glaucoma diagnosis. 

\paragraph{Evidence-based Reasoning Paradigm}
During sequential reasoning, the system evaluates the output $r_i=<k_{r_i},v_{r_i}>$ at each step $\mathcal{P}'_i$ to determine a status $s_i \in \{\text{Continue}, \text{Terminate}, \text{Complete}\}$, which guides whether the reasoning process should proceed. $s_i$ is determined by:
\begin{equation}
s_i =
\begin{cases}
\text{Complete}, & \text{if } k_{r_i} \in \mathcal{I} \\
\text{Terminate}, & \text{if } k_{r_i} \notin \mathcal{I} \land \neg \phi(v_{r_i}) \\
\text{Continue}, & \text{if } k_{r_i} \notin \mathcal{I} \land \phi(v_{r_i})
\end{cases}
\end{equation}
Here, $\phi$ is a state assessment function implemented by the VLM, which evaluates the reliability of the result data $v_{r_i}$ based on the quality of input data $v_{o_i}$, and the plausibility of $v_{r_i}$. 
If $s_i = \text{Continue}$, the output $r_i$ is treated as evidence $e$ and used as the input $o_{i+1}$ for the next reasoning step. If $s_i = \text{Terminate}$, the process is halted, as $r_i$ is deemed unreliable and may hinder subsequent reasoning and lead to incorrect diagnoses.
This process continues iteratively until $s_i=Complete$.
The final output is a set of reasoning results of indicators: $\mathcal{R}_{final}=\{r_i|s_i=\text{Complete\}}$.
The VLM then assigns risk-based weights $\mathcal{W}$ to $\mathcal{R}_{final}$ to balance different indicators' results, guided by the clinical guideline $\mathcal{G}$. The final risk score $\rho$ is computed as a weighted sum:
\begin{equation}
    \rho = \sum_{i=0}^{|\mathcal{R}_{final}|=l} w_i v_{r_i}, s.t. w_i \in W, r_i\in \mathcal{R}_{final}
\end{equation}
The final diagnosis is then made by comparing $s$ with a risk threshold $\theta$. Through evidence-based reasoning, MedAgent-Pro integrates reliable external evidence with expert knowledge to improve diagnostic decision-making, thus promoting a modern diagnosis workflow.


\begin{table*}[!t]
\centering
\renewcommand{\arraystretch}{1.1}
\caption{Comparison with general VLMs and medical agentic systems on REFUGE2, MITEA and NEJM datasets (\%). "Opht." is the short form of Ophthalmology. }
\label{tab:sota}
\resizebox{0.98\textwidth}{!}{
\begin{tabular}{c|cc|cc|c|c|c|c|c}
\hline
\multirow{2}{*}{Method} & \multicolumn{2}{c|}{Glaucoma} & \multicolumn{2}{c|}{Heart Disease} & \multicolumn{5}{c}{NEJM (Acc)}\\ \cline{2-10}
 & bAcc & F1 & bAcc & F1 & All & Cell Imaging & Chest X-ray & CT \& MRI & Opht. Imaging\\
\hline
GPT-4o \cite{gpt4}  & 56.4 & 21.1 & 56.8 & 28.1 & 70.9 & 74.6 & 54.5 & 63.4 & 70.5\\
Janus-Pro-7B \cite{janus}  & 53.4 & 13.3 & 52.3 & 10.7 & 30.0 & 32.2 & 36.4 & 29.3 & 28.2  \\
LLaVA-Med \cite{llavamed} & 50.0  & 0.0 & 50.0  & 0.0 & 26.2 & 39.0 & 18.2 & 24.4 & 35.9\\
BioMedClip \cite{biomedclip} & \underline{58.1} & 21.3 & 47.0 & \underline{37.8} & 27.9 & 27.1 & 40.0 & 29.3 & 20.5\\
Qwen2.5-7B-VL \cite{Qwen} & 54.3 & 16.3 & 50.0 & 0.0 & 41.8 & 54.2 & 30.9 & 36.6 & 44.9 \\
InternVL2.5-8B \cite{internvl} &51.8  & 13.8 & 49.7  & 3.6 & 42.2 & 37.3  & 40.0 & 41.5 & 35.9 \\
\hline
MedAgents \cite{medagents} (ACL'24) & 52.1 & 8.9 & 51.1 &15.9 & 66.1 & 69.9 & 51.5 & 58.7 &68.2\\
MMedAgent \cite{mmedagent} (EMNLP'24) & 52.4  & 16.3  & 55.0 & 26.7 & 71.7 & 73.8 & \underline{56.4} & 65.3  & 70.5\\
MDAgent \cite{mdagent}  (NeurIPS'24)& 56.8 & \underline{22.2} & \underline{57.2} & 30.3 & \underline{73.8} & \underline{79.6} & 52.9 & \underline{67.3} & \underline{73.0}\\
MedAgent-Pro (Ours) & \textbf{90.4} & \textbf{76.4} & \textbf{77.8} & \textbf{72.3} & \textbf{81.7} & \textbf{90.5} & \textbf{69.1} & \textbf{72.7} & \textbf{89.7}\\
\hline
\end{tabular}}
\vspace{-5pt}
\end{table*}

\begin{table*}[!t]
\renewcommand{\arraystretch}{1.1}
\centering
\caption{Comparison with general VLMs on the MIMIC dataset (\%). “Avg.” is the average performance for the 12 sub-tasks. Only bAcc values are presented due to space limitations.}
\label{tab:mimic}
\resizebox{0.98\textwidth}{!}{%
\begin{tabular}{c|c|c c c c c c}
\hline
Method & Avg. & Atelectasis & Cardiomegaly & Consolidation & Edema & \multicolumn{1}{c}{\begin{tabular}[c]{@{}c@{}}Enlarged\\Cardiomediastinum\end{tabular}} & Fracture \\
\hline
GPT-4o \cite{gpt4}      & \underline{58.3}   & 68.7 & \underline{64.3} & 60.5 & \textbf{61.2} & 53.2 & \underline{56.3}   \\
Janus-Pro-7B \cite{janus}   & 51.9   & 61.7 & 54.1 & 45.2 & 52.5 & 34.1 & 50.0 \\
LLaVA-Med \cite{llavamed}      & 50.2   & 50.0 & 51.5 & 50.0 & 50.0 & 50.0 & 50.0 \\
BioMedClip \cite{biomedclip}     & 57.9   & 48.6 & 62.7 & \underline{62.2} & 50.0 & \underline{61.1} & 50.0 \\
Qwen2.5-7B-VL \cite{Qwen}  & 55.2   & \underline{69.6} & 58.1 & 54.5 & 48.8 & 50.8 & 50.0 \\
InternVL2.5-8B \cite{internvl} & 51.6   & 57.7 & 48.6 & 51.7 & 47.5 & 50.8 & 43.8 \\ \hline
MedAgent-Pro   & \textbf{72.0}   & \textbf{85.5} & \textbf{74.2} & \textbf{66.3} & \underline{59.2} & \textbf{75.4}    & \textbf{68.5}   \\
\hline
Method & Avg. & \multicolumn{1}{c}{\begin{tabular}[c]{@{}c@{}}Lung\\Leision\end{tabular}} & \multicolumn{1}{c}{\begin{tabular}[c]{@{}c@{}}Lung\\Opacity\end{tabular}} & \multicolumn{1}{c}{\begin{tabular}[c]{@{}c@{}}Pleural\\Effusion\end{tabular}} & Pneumonia & Pneumothorax & \multicolumn{1}{c}{\begin{tabular}[c]{@{}c@{}}Supporting\\Devices\end{tabular}} \\
\hline
GPT-4o \cite{gpt4}         & \underline{58.3}  & 38.6 & 63.1 & 62.7 & \underline{59.6} & 47.7 & 63.4   \\
Janus-Pro-7B \cite{janus}  & 51.9  & 59.1 & 42.1 & 57.4 & 57.4 & 47.2 & 62.4 \\
LLaVA-Med \cite{llavamed}      & 50.2  & 50.0 & 50.0 & 50.0 & 50.1 & 50.5 & 50.0 \\
BioMedClip \cite{biomedclip}     & 57.9  & \textbf{81.8} & 50.0 & \underline{64.7} & 55.1 & \underline{53.0} & 55.4 \\
Qwen2.5-7B-VL \cite{Qwen}  & 55.2  & 43.2 & \underline{64.4} & 48.2 & 59.3 & 34.9 & \underline{80.8} \\
InternVL2.5-8B \cite{internvl} & 51.6  & 59.1 & 49.1 & 54.2 & 54.0 & 47.6 & 55.3 \\ \hline
MedAgent-Pro   & \textbf{72.0}     & \underline{72.2} & \textbf{67.6}   & \textbf{77.8}    & \textbf{62.1}    & \textbf{65.6}    & \textbf{89.2}   \\
\hline
\end{tabular}%
}
\vspace{-5pt}
\end{table*}
 
\section{Experiment}
\vspace{-5pt}
We compare our method with VLMs, task-specific models and medical agentic systems in Section~\ref{sec:com}. Ablation studies and in-depth analyses are presented in Section~\ref{sec:abla} to demonstrate the effectiveness of our approach. To assess clinical relevance, we also conduct human evaluation in Section~\ref{sec:human}. 
\vspace{-5pt}
\subsection{Experimental Setup}
\vspace{-5pt}
\paragraph{Dataset} We conduct experiments on four datasets with increasingly challenging settings. The REFUGE2 dataset \cite{refuge2} is used for glaucoma diagnosis, and the MITEA dataset \cite{mitea} for heart disease diagnosis (e.g., dilated cardiomyopathy, amyloidosis), both of which are suited for evaluating in-depth diagnostic reasoning.
To assess multi-disease diagnosis for individual patients, we sample 442 chest X-ray cases from 100 patients from the MIMIC dataset \cite{mimic}, each case involving the identification of up to 12 potential thoracic conditions or abnormalities.
We further employ New England Journal of Medicine (NEJM) database \cite{web:nejm}, where we compile 992 real-world diagnostic cases, encompassing over 10 anatomical regions, 10 imaging modalities, and 50 diseases.
For cases involving cell or ophthalmology imaging, visual tools are available. However, non-clinical images (e.g., everyday photographs) often lack compatible tool support.
\vspace{-5pt}
\paragraph{Evaluation Metrics} Three metrics are used to evaluate performance: for REFUGE2, MITEA, and MIMIC datasets, we report balanced accuracy (bAcc) and the F1 score. For the NEJM dataset, where tasks are framed as multiple-choice questions in accordance with its evaluation protocol, we report the accuracy rate. The best results are highlighted in \textbf{bold}, and the second-best are \underline{underlined}.
\vspace{-5pt}
\paragraph{Implementation Detail} During the MedAgent-Pro workflow, we use GPT-4o \cite{gpt4} as the baseline VLM, and implement the RAG agent using LangChain \cite{langchain}. For fair comparison, all baseline medical agentic systems also adopt GPT-4o as their underlying VLM.

\begin{table}[th]
  \centering
      \vspace{-10pt}
  \renewcommand{\arraystretch}{1.1}
  \caption{Comparison with task-specific models (\%).}     
  \resizebox{0.98\textwidth}{!}{%

    \begin{tabular}{c|c|c|cc|c|c}
    \hline 
    \multicolumn{2}{c|}{REFUGE2 winners} & \multicolumn{3}{c|}{Ophthalmology VLMs} & \multicolumn{2}{c}{Chest X-Ray VLMs}\\ \hline
    Team Name & AUC & Method & bAcc & F1 & Method & bAcc\\
    \hline
    VUNO EYE TEAM & \underline{88.3}   & RetiZero \cite{retizero}& 50.8 & 18.4 & Maira-2 \cite{maria} & 64.1\\
    MIG & 87.6 & VisionUnite \cite{visionunite} & \underline{85.8} & \underline{73.1} & CheXagent \cite{chexagent} & \underline{69.1}\\
    \hline
    MedAgent-Pro & \textbf{95.1} & MedAgent-Pro & \textbf{90.4} & \textbf{76.4} & MedAgent-Pro & \textbf{72.0}\\
    \hline
    \end{tabular}}
    \vspace{-10pt}
  \label{tab:refuge_winner}%
 \end{table}%
 
\subsection{Comparision Experiments}
\label{sec:com}
\vspace{-5pt}
\paragraph{Comparison with General VLMs}
We have included comparisons with advanced VLMs such as BioMedClip \cite{biomedclip}, GPT-4o \cite{gpt4}, LLaVA-Med \cite{llavamed}, Janus \cite{janus}, Qwen \cite{Qwen}, and InternVL \cite{internvl} on all settings. 
As the evaluated VLMs are not designed to process 3D images, we randomly select three slices from the 3D echocardiography in the MITEA dataset as visual input. This process is repeated ten times and the mean performance is reported to mitigate sampling variability.
As presented in Table \ref{tab:sota} and \ref{tab:mimic}, the proposed MedAgent-Pro framework significantly outperforms existing VLMs across all datasets. 
Specifically, it achieves improvements in bAcc of 34.0\% and 21.0\%, and gains in F1 score of 55.3\% and 44.2\% for glaucoma and heart disease diagnosis respectively, compared to GPT-4o. 
These results highlight the effectiveness of our MedAgent-Pro workflow in handling complex diagnostic tasks, particularly those requiring quantitative indicators such as cup-to-disc ratio or left ventricular ejection fraction. 
By integrating visual tools directly into the reasoning process, MedAgent-Pro enables precise indicator calculation and enhances diagnostic performance, effectively addressing the existing limitations of VLMs. 

Across diverse real-world diagnostic scenarios in the NEJM database, MedAgent-Pro demonstrates significant performance gains in domains with visual tool support, such as cell imaging, chest X-rays, CT/MRI, and ophthalmology. Furthermore, it maintains strong performance in cases without visual tool support, achieving an overall improvement of 7.9\%, which demonstrates its strong robustness and generalizability across a variety of diagnostic tasks.
In addition, as shown in table \ref{tab:mimic}, chest X-ray diagnosis involves certain tasks like \textit{Cardiomegaly} which rely on precise quantitative measurements like the cardiothoracic ratio. Meanwhile, others such as \textit{Fracture} detection, require detailed step-by-step analysis to identify subtle abnormalities.
MedAgent-Pro achieves leading performance across most tasks, with an average performance gain of 13.7\%.

\vspace{-5pt}
\paragraph{Comparison with Medical Agentic Systems}
We also compare MedAgent-Pro with advanced medical agentic frameworks, including MedAgents \cite{medagents}, MMedAgent \cite{mmedagent} and MDAgent \cite{mdagent}.  Since MedAgents and MDAgent are originally designed for text-based questions, we adapt them into a VQA setting with their core mechanism.
As shown in Table \ref{tab:sota}, MedAgent-Pro consistently outperforms these methods across all diseases and domains. This performance advantage stems from the fact that prior methods are primarily designed for basic question answering or as modular toolboxes, lacking the capacity to handle complex, multi-modal clinical scenarios. In contrast, MedAgent-Pro incorporates retrieval-based diagnostic steps and seamlessly integrates visual tools into its reasoning process, enabling effective and comprehensive decision-making support in clinical applications.

\vspace{-5pt}
\paragraph{Comparison with Task-specific Models}
Additionally, we compare MedAgent-Pro with SOTA task-specific methods, i.e., for glaucoma diagnosis \cite{retizero, visionunite} and chest X-ray analysis \cite{maria,chexagent}. For glaucoma diagnosis, we also compare with the winners from the REFUGE2 challenge leaderboard \cite{refuge2}. As the leaderboard only reports the AUC metric, our comparison is limited to this metric.

As shown in Table \ref{tab:refuge_winner}, MedAgent-Pro outperforms these task-specific methods, despite the VLMs in MedAgent-Pro remaining zero-shot.
In glaucoma diagnosis, the AUC metric has improved by 6.8\%, while the bAcc and F1 scores have increased by 4.6\% and 3.3\%, respectively. 
This finding further demonstrates that integrating specialized tools with general VLMs can achieve performance comparable to domain-specific models, emphasizing the potential of the MedAgent-Pro framework.

\begin{table}[ht]
  \centering
  \renewcommand{\arraystretch}{1.1}
  \begin{minipage}[t]{0.65\textwidth}
    \centering
    \captionof{table}{Ablation on key components, including Planning, Evidence-based Reasoning and Quantitative Analysis.}
    \label{tab:ablation}
      \resizebox{1\textwidth}{!}{%
    \begin{tabular}{ccc|cc|cc}
      \hline
      \multicolumn{3}{c|}{Setting} & \multicolumn{2}{c|}{Glaucoma} & \multicolumn{2}{c}{Heart Disease}\\ 
      \hline
      Planning & Quan. & Evi. Reasoning & bAcc & F1 & bAcc & F1\\
      \hline
      &         &              & 56.4  & 21.1 & 56.8 & 28.1\\
      \checkmark &  &             & 75.9  & 36.5 & 63.3 & 45.9\\
      \checkmark &  \checkmark &  & \underline{88.5} & \underline{71.0} & \underline{73.4} & \underline{66.6}\\
      \checkmark & \checkmark & \checkmark  & \textbf{90.4} & \textbf{76.4} & \textbf{77.8} & \textbf{72.3}\\
      \hline
    \end{tabular}}
    \vspace{-10pt}
  \end{minipage}%
  \hfill
  \begin{minipage}[t]{0.3\textwidth}
    \vspace{10pt}                  
    \centering
    \captionof{table}{Ablation on qualitative indicators analysis.}
    \label{tab:able_quali}
     \resizebox{1\textwidth}{!}{
    \begin{tabular}{c|cc}
      \hline
      Method      & bAcc & F1   \\ 
      \hline
      GPT-4o      & 90.4 & 76.4 \\ 
      VisionUnite & 92.9 & 79.1 \\ 
      \hline
    \end{tabular}}
    \vspace{-10pt}
  \end{minipage}
\end{table}

\subsection{Ablation Study and Detailed Analysis}
\label{sec:abla}
\paragraph{Effectiveness of the Proposed Key Components}
We conduct an ablation study on glaucoma and heart disease diagnosis to evaluate the effectiveness of three key modules: planning, quantitative analysis, and evidence-based reasoning, while each module builds upon the previous one.
As shown in Table~\ref{tab:ablation}, incorporating planning significantly improves overall performance, while integrating visual tools for quantitative analysis brings further gains, with F1 scores increasing by 34.5\% and 20.7\%, respectively. The addition of evidence-based reasoning further enhances the consistency of plan execution and the reliability of analysis, thereby reaching the best performance.
These results validate the effectiveness of the proposed components and demonstrate their complementary roles in jointly enabling truly evidence-based medical reasoning.

\begin{figure}[ht]
    \centering
    \begin{minipage}[b]{0.48\linewidth}
        \centering
        \includegraphics[width=\linewidth]{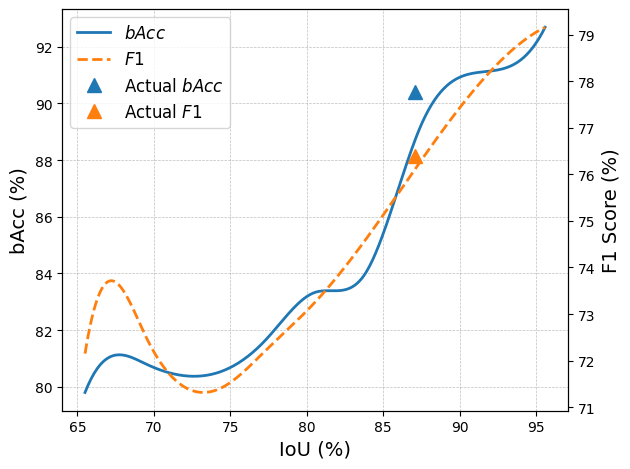}
        \caption{Ablation on quantitative indicator analysis that reveals how segmentation accuracy influences diagnostic outcomes.}
        \label{fig:noisy_label}
        \vspace{-10pt}
    \end{minipage}%
    \hfill
    \begin{minipage}[b]{0.48\linewidth}
        \centering
        \includegraphics[width=\linewidth]{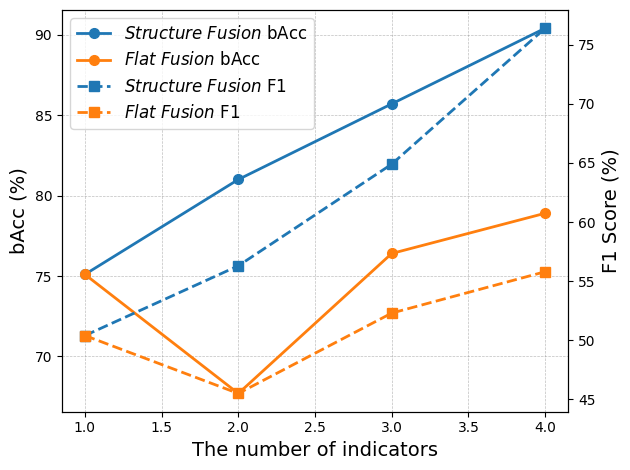}
        \caption{Comparison of two kinds of decision-making ways on Glaucoma diagnosis under different indicator numbers.}
        \label{fig:decision-making}
        \vspace{-10pt}
    \end{minipage}
\end{figure}

\paragraph{Analysis of the Impact of Indicator Accuracy}
Both qualitative and quantitative analyses at each reasoning step may introduce errors. To assess their respective impact on final diagnostic accuracy, we conduct ablation studies on glaucoma diagnosis. Since the accuracy of qualitative analysis is difficult to quantify, we use an ophthalmic-specific model VisionUnite \cite{visionunite} to conduct qualitative analysis in our workflow instead of the original GPT-4o.
As shown in Table \ref{tab:able_quali}, the results show only marginal improvement, indicating that general-purpose VLMs, such as GPT-4o, when guided by medical guidelines, are sufficient for qualitative analysis without requiring additional domain-specific tools. 
In addition, in Fig.~\ref{fig:noisy_label}, we simulate noisy segmentation masks following prior works \cite{noisy1,noisy2,noisy3,noisy4}, and observe that higher segmentation accuracy consistently yields better diagnostic performance. These findings suggest that in multi-modal diagnosis, evidence-based quantitative analysis plays a more critical role than experience-driven qualitative assessment.

\vspace{-5pt}
\paragraph{Analysis of Decision-Making Strategies}
We further explore alternative decision-making strategies to integrate clinical indicators. In addition to our proposed \textit{structured fusion}, which assigns risk-based weights to clinical indicators, we evaluate \textit{flat fusion}, where all raw indicators are directly fed into the VLM for final decision-making.
As shown in Fig.~\ref{fig:decision-making}, structured fusion consistently outperforms flat fusion across varying indicator counts by assigning explicit weights, leading to more balanced and comprehensive decisions, whereas VLMs often focus on partial cues.

\vspace{-5pt}
\section{Human Evaluation with Clinical Experts}
\label{sec:human}
\begin{figure}[ht]
    \centering
    \includegraphics[width=0.9\linewidth]{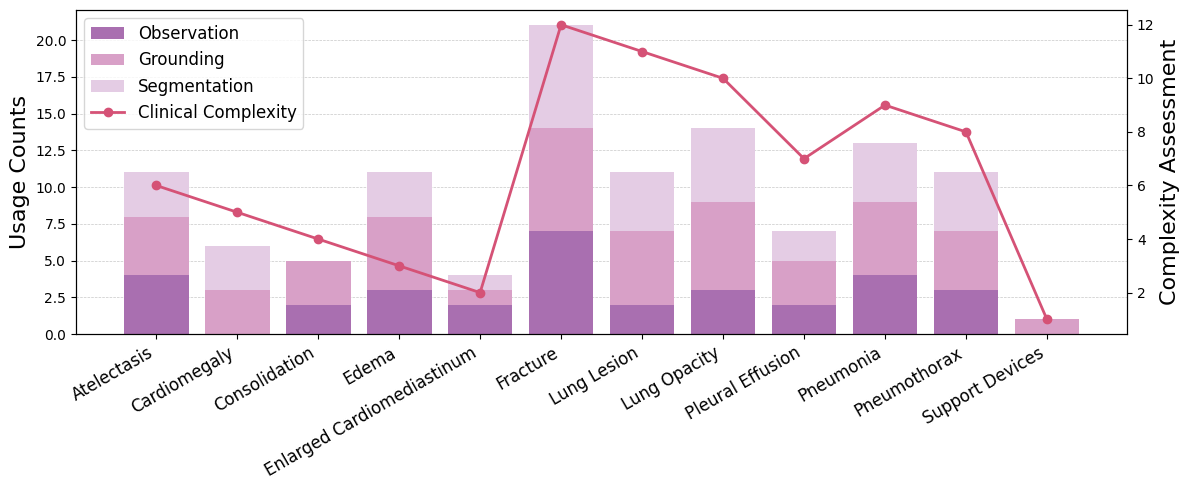}
    \caption{Diagnostic plan complexity vs. clinician assessment by subtasks on MIMIC.}
    \label{fig:tool_usage}
    \vspace{-12pt}
\end{figure}
\paragraph{Alignment with Real-World Clinical Workflow}
To assess how well MedAgent-Pro reflects real clinical workflows, we quantified the number of steps it performs across 12 chest X-ray diagnostic tasks.
We compared the results with thoracic clinicians’ rankings, where each task was rated from 1 to 12 based on perceived diagnostic difficulty and time demand.
As shown in Fig. \ref{fig:tool_usage}, most tasks show a clear positive correlation between total step count and physician‐rated complexity. For instance, \textit{Fracture} is the most complex and involves the highest number of steps, while \textit{Support Devices} require the fewest step and is ranked least complex.  

Notably, conditions like \textit{Pleural Effusion} and \textit{Cardiomegaly} benefit from visual tool integration, significantly reducing workflow steps. In contrast, tasks such as \textit{Fracture} and \textit{Edema} remain step-intensive. This is because diagnoses relying on quantitative indicators (e.g., the cardiothoracic ratio) can be automated by visual tools, while those requiring qualitative assessment still depend on sequential reasoning.
The findings demonstrate the effectiveness and practical compatibility of MedAgent-Pro's structured, evidence-based workflow with real-world clinical diagnostic process.
    
\vspace{-5pt}

\paragraph{Assessment of Generated Diagnostic Content Across Other Methods}


To further assess diagnostic quality beyond accuracy, we conduct validation with clinical experts on glaucoma and chest X-ray diagnosis.
The clinicians rate the diagnostic outputs from both VLMs and MedAgent-Pro across five dimensions—relevance, comprehensiveness, clinical reliability, reasoning coherence, and language clarity—using a 1 to 5 scale following \cite{he1,he2,he3,he4,he5}. 
MedAgent-Pro outperforms other VLMs across five dimensions and demonstrates strong stability across diverse cases, underscoring the alignment of our structured, evidence-based approach with modern medical diagnosis standards.
\begin{figure}[ht]
    \centering
    \includegraphics[width=1\linewidth]{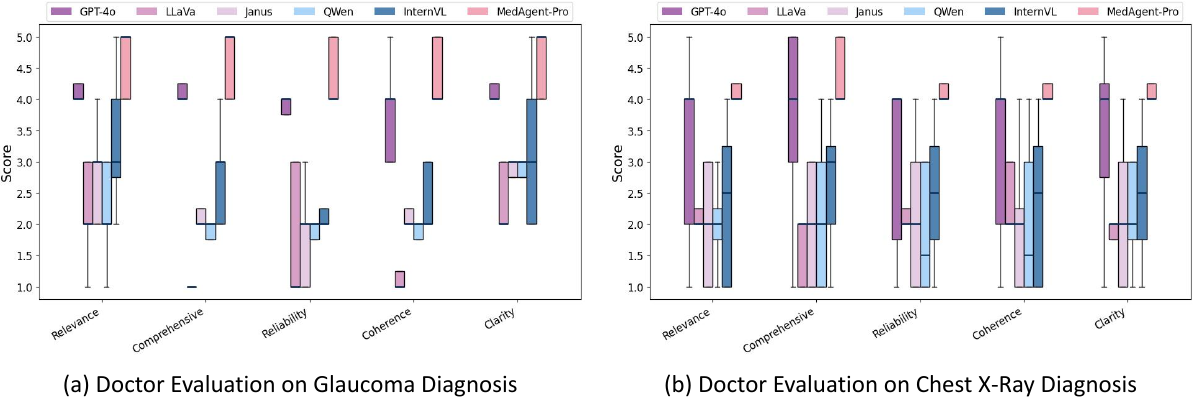}
    \caption{Evaluation on glaucoma and chest X-ray diagnosis by clinical experts.}
    \label{fig:human_evaluation}
    \vspace{-13pt}
\end{figure}

\section{Conclusion}
\paragraph{Broader Impact} This paper introduces MedAgent-Pro, a reasoning-based agentic system designed to deliver accurate, multi-modal medical diagnoses, addressing the limitations of treating diagnosis as an empirical task. MedAgent-Pro incorporates medical guidelines for planning, integrating quantitative analysis, and verifying the reliability of each reasoning step. Our method bridges the gap between AI systems and clinical procedures. This represents a significant step toward the core principles of evidence-based medicine and advancing the practical application of AI in healthcare.

\paragraph{Limitations} While advancing automated clinical diagnosis, this work still has several limitations.
The proposed framework depends on the availability of visual tools, which remain limited in certain medical domains. In addition, qualitative analysis still relies on VLMs, which are prone to VLMs' inherent inconsistency and hallucination.
Addressing these limitations will be essential to further improve the reliability and clinical impact of computer-aided diagnosis.

\section{Acknowledgment}
This work was supported by Ministry of Education Tier 1 Start up grant, NUS, Singapore (A-8001267-01-00); Ministry of Education Tier 1 grant, NUS, Singapore (A-8003261-00-00). Junde Wu is supported by the Engineering and Physical Sciences Research Council (EPSRC) under grant EP/S024093/1 and GE HealthCare.



\end{document}